\def\paperTitle{Transfer Your Perspective:\\
Controllable 3D Generation from Any Viewpoint in a Driving Scene
}

\def\authorBlock{
Tai-Yu Pan\textsuperscript{1}, Sooyoung Jeon\textsuperscript{1}, Mengdi Fan\textsuperscript{1}, Jinsu Yoo\textsuperscript{1}, Zhenyang Feng\textsuperscript{1},\\
Mark Campbell\textsuperscript{2}, Kilian Q. Weinberger\textsuperscript{2}, Bharath Hariharan\textsuperscript{2}, Wei-Lun Chao\textsuperscript{1}
\\
\vspace{-5pt}
\\
\textsuperscript{1}The Ohio State University, \textsuperscript{2}Cornell University
}

\newif\ifreview 
\newif\ifarxiv \newcommand{\arxiv}{\arxivtrue}
\newif\ifcamera 
\newif\ifrebuttal 

\arxiv

\pdfoutput=1
\documentclass[10pt,twocolumn,letterpaper]{article}
\ifreview \usepackage[review]{cvpr} \fi
\ifarxiv \usepackage[pagenumbers]{cvpr} \fi
\ifrebuttal \usepackage[rebuttal]{cvpr} \fi
\ifcamera \usepackage{cvpr} \fi

\usepackage{graphicx}	
\usepackage{amsmath}	
\usepackage{amssymb}	
\usepackage{booktabs}
\usepackage{times}
\usepackage{microtype}
\usepackage{epsfig}
\usepackage{caption}
\usepackage{float}
\usepackage{placeins}
\usepackage{color, colortbl}
\usepackage{stfloats}
\usepackage{enumitem}
\usepackage{tabularx}
\usepackage{xstring}
\usepackage{multirow}
\usepackage{xspace}
\usepackage{url}
\usepackage{subcaption}
\usepackage{xcolor}
\usepackage[hang,flushmargin]{footmisc}

\usepackage{adjustbox}
\usepackage{arydshln}

\ifcamera \usepackage[accsupp]{axessibility} \fi

\definecolor{customgreen}{RGB}{0, 98, 65}
\definecolor{aliceblue}{rgb}{0.94, 0.97, 1.0}
\definecolor{beaublue}{rgb}{0.74, 0.83, 0.9}
\definecolor{lightcyan}{rgb}{0.88, 1.0, 1.0}

\newcommand{\ours}{\method{TYP}\xspace}

\newcommand{\nbf}[1]{{\noindent \textbf{#1.}}}

\newcommand{\supp}{supplemental material\xspace}
\ifarxiv \renewcommand{\supp}{appendix\xspace} \fi

\newcommand{\R}[1]{{%
    \textbf{%
        \ifstrequal{#1}{1}{\textcolor{magenta}{mo5W}}{%
        \ifstrequal{#1}{2}{\textcolor{teal}{Wi7E}}{%
                           \textcolor{cyan}{HSRB}%
        }}%
    }%
}}

\usepackage{xr-hyper}

\makeatletter
\newcommand*{\addFileDependency}[1]{
  \typeout{(#1)}
  \@addtofilelist{#1}
  \IfFileExists{#1}{}{\typeout{No file #1.}}
}

\makeatother
\newcommand*{\myexternaldocument}[1]{
    \externaldocument{#1}
    \addFileDependency{#1.tex}
    \addFileDependency{#1.aux}
}

\definecolor{cvprblue}{rgb}{0.21,0.49,0.74}
\usepackage[pagebackref,breaklinks,colorlinks,allcolors=cvprblue,citecolor=citecolor,linkcolor=linkcolor]{hyperref}
\usepackage{xr}
\usepackage[capitalize]{cleveref}
\crefname{section}{Sec.}{Secs.}
\crefname{table}{Table}{Tables}
\crefname{figure}{Fig.}{Figs.}

\ifarxiv \crefname{appendix}{App.}{Apps.}
\else \crefname{appendix}{Suppl.}{Suppls.} \fi

\usepackage{amsmath,amsfonts,bm}

\definecolor{citecolor}{rgb}{0,0.08,0.45}
\definecolor{linkcolor}{RGB}{187,18,26}

\def\eqref#1{equation~\ref{#1}}

\def\1{\bm{1}}

\def\rmF{{\mathbf{F}}}

\def\vx{{\bm{x}}}
\def\vy{{\bm{y}}}

\DeclareMathAlphabet{\mathsfit}{\encodingdefault}{\sfdefault}{m}{sl}
\SetMathAlphabet{\mathsfit}{bold}{\encodingdefault}{\sfdefault}{bx}{n}

\def\gN{{\mathcal{N}}}

\def\sE{{\mathbb{E}}}

\def\sR{{\mathbb{R}}}

\newcommand{\Ls}{\mathcal{L}}

\newcommand{\method}[1]{\textsc{#1}}

\usepackage{cuted}

\unless\ifarxiv \myexternaldocument{_supplementary} \fi

\begin{document}
\title{\paperTitle}
\author{\authorBlock}
\maketitle

\begin{strip}
\vspace{-3.3em}
\setlength{\linewidth}{\textwidth}
\setlength{\hsize}{\textwidth}
\centering
\includegraphics[trim={0cm, 0cm, 0cm, 0cm},clip,width=\textwidth]{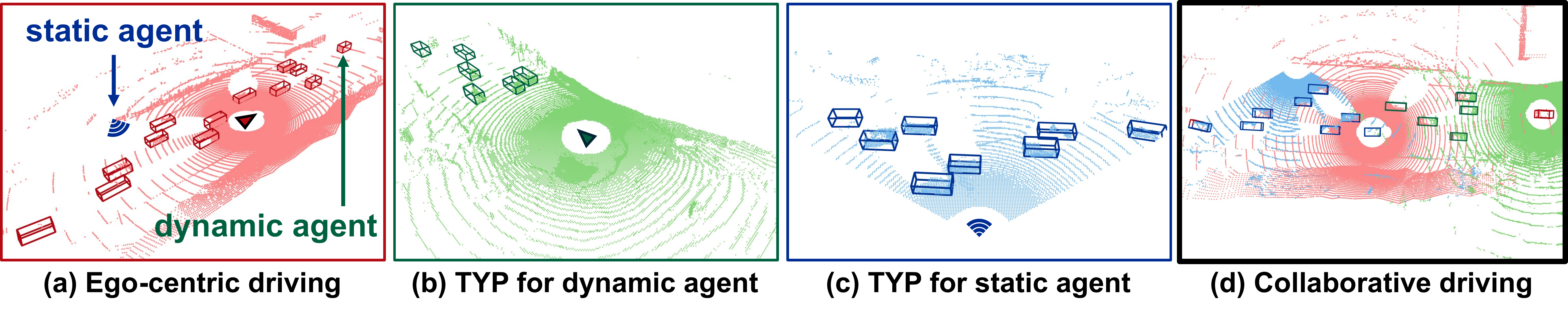}
\vspace{-8mm}
\captionof{figure}{
\label{fig:teaser}
\small \textbf{Illustration of the proposed problem and solution, Transfer Your Perspective (\ours).} (a) A given sensory data captured by the {\color{red} ego-car (red triangle)}. (b) A generated sensory data by \ours, seeing from the viewpoint of \textcolor{customgreen}{another vehicle (green triangle)} in the same scene. (c) A generated sensory data, seeing from an imaginary {\color{blue} static agent like roadside units (blue icon)}. (d) Putting all the sensory data together, given or generated, \ours enables the development of collaborative perception with little or no real collaborative driving data.}
\end{strip}

\begin{abstract}

Self-driving cars relying solely on ego-centric perception face limitations in sensing, often failing to detect occluded, faraway objects. Collaborative autonomous driving (CAV) seems like a promising direction, but collecting data for development is non-trivial. It requires placing multiple sensor-equipped agents in a real-world driving scene, simultaneously! As such, existing datasets are limited in locations and agents. We introduce a novel surrogate to the rescue, which is to generate realistic perception from different viewpoints in a driving scene, conditioned on a real-world sample---the ego-car's sensory data. This surrogate has huge potential: it could potentially turn any ego-car dataset into a collaborative driving one to scale up the development of CAV.
We present the very first solution, using a combination of simulated collaborative data and real ego-car data. 
Our method \textbf{Transfer Your Perspective (\ours)} learns a conditioned diffusion model whose output samples are not only realistic but also consistent in both semantics and layouts with the given ego-car data. Empirical results demonstrate \ours's effectiveness in aiding in a CAV setting. In particular, \ours enables us to (pre-)train collaborative perception algorithms like early and late fusion with little or no real-world collaborative data, greatly facilitating downstream CAV applications.

\end{abstract}

\section{Introduction}
\label{sec:intro}

Seeing the world from an ego-centric perspective, a self-driving car risks being ``narrow-sighted,'' limiting its ability to respond appropriately in dynamic driving environments. For instance, it should slow down or honk if a pedestrian is about to cross the road or if another vehicle is merging into its lane. However, these actions depend on the car’s ability to detect traffic participants, which may be occluded by a large bus or a building at a sharp intersection. %
This occlusion problem can hardly be addressed by mounting a few more sensors on the car. We argue that a self-driving car must move beyond its ego-centric perspective.

One intuitive way is to collaborate with nearby ``agents'' that observe the scene from different angles, such as other sensor-equipped cars or static devices like roadside units (RSUs). When trained together to address GPS errors and synchronization delays~\citep{yoo2024learning}, collaborative perception has been shown to significantly improve the accuracy of each agent's perception, particularly in detecting occluded or distant objects~\citep{xu2022opv2v,chen2019cooper,rawashdeh2018collaborative, chen2019fcooper,wang2020v2vnet,liu2020when2com,li2021learning,hu2022where2comm,xu2022v2xvit,xu2023v2v4real,xiang2024v2x}. 
\emph{However, collecting training data for collaborative perception is never easy.} 
Unlike single-agent data, which can be collected by simply driving a car on the road, collaborative data requires the simultaneous presence of multiple agents in the same driving scene. For dynamic agents like vehicles, precise coordination is needed to ensure they are within communication range. These challenges limit existing works in scale and the number of agents (typically just two). While one may leverage simulated data by game engines~\citep{Dosovitskiy17carla}, they often fail to capture the diversity of real-world scenes. %
We thus ask:
\vspace{-2mm}
\begin{center}
    \color{blue}
    \emph{
    Can we obtain realistic data for learning collaborative perception with much less effort, ideally as easily as single-agent data?
    }
\end{center}
\vspace{-2mm}
Specifically, given the vast amount of single-agent LiDAR data already collected across various driving environments, is it possible to convert each of them into collaborative perception data by generating additional point clouds from alternative reference viewpoints within the same scene?

At first glance, \textbf{this question may seem overly ambitious for three reasons.} First, for collaborative perception to be effective, the generated point cloud must provide information that the ego-car’s point cloud cannot capture, such as occluded surfaces invisible to the ego-car. This creates a chicken-and-egg problem: if the ego-car’s point cloud is all we initially know about the scene, how can we deduce any additional information?
Second, for training purposes, the generated data must be realistic. It should replicate the constraints and data patterns of a real sensor as if it were positioned at the reference viewpoint, producing no points in occluded areas and fewer points in distant areas. 
Last but not least, in regions commonly perceivable from both the reference and ego-car's viewpoints, the generated data should align in layouts and semantics with the ego-car's data. On the surface, this may seem as simple as copying the ego-car's data, but doing so would violate the second requirement. In essence, the generated data should resemble a point cloud seen from the reference viewpoint, not the ego-car.

That said, after a deeper look at the question, \textbf{we argue that it is achievable with three key insights.} First, given semantic information around the ego-car for conditioning, existing research has shown promising progress in generating realistic point clouds~\citep{ran2024towards, hu2025rangeldm, wu2024text2lidar, nakashima2024lidar-r2dm, zyrianov2024lidardm, xiong2023ultralidar, zyrianov2022learning-lidargen}. This is especially encouraging, as most existing single-agent datasets provide 3D object labels, and they can be translated to obtain semantic information centered around the reference position. This further implies that we can generate occluded object surfaces as seen from the ego-car if they are visible from the reference viewpoint. Second, semantic information is easily editable: even if the translated object map has noticeable empty areas due to limited ego-car perception, object boxes can be manually added to make the map appear more realistic from the reference viewpoint. Lastly, in commonly perceivable regions where the generated data needs to meet two physical constraints, we could leverage simulators based on computational graphics and optics. Specifically, we can use simulated data from the two views to train a conditioned generative model that maps one viewpoint to the other.

Building upon these insights, we propose \textbf{Transfer Your Perspective (\ours)}, a novel research problem and the very first solution for \emph{generating a realistic point cloud from any viewpoint in a scene, given real ego-car's point cloud and semantic labels as conditions.} \ours assumes access to \textbf{1)} a simulated collaborative driving dataset with multiple agents perceiving the same scenes from different viewpoints and \textbf{2)} a real single-agent driving dataset; both are labeled.
Our solution involves a \textbf{conditioned latent diffusion model}~\citep{rombach2022high} and a \textbf{dedicated two-stage training process}. In the first stage, we consider a single-agent scenario and train the model using real data conditioning only on object locations. This equips the model with the ability to generate diverse and realistic scenes. We denote the learned model by $P(\vx|\vy)$, where $\vy$ stands for the semantic condition and $\vx$ for the point cloud.  
In the second stage, we incorporate the simulated data to learn \emph{how to ground generation on data from another agent's viewpoint} so that the model can produce a semantically consistent reference point cloud given the ego-car's data. 
We learn a lightweight conditioning module to turn $P(\vx|\vy)$ into $P(\vx_r|\vx_e, \vy_r)$, where $e$ and $r$ indicate ego and reference views, respectively. We note that $\vx_e$ was pre-translated to center around the reference position. 

One challenge in \ours is the domain gap between real and simulated data. Due to differences in sensor configurations and placements, data collection environments, and the sim-to-real gap, the two sets of point clouds inevitably exhibit discrepant distributions, patterns, and densities. To address this, \emph{we insert a domain adaptation step between the two training stages}. We train a separate encoder-decoder for the simulated data while enforcing a constraint to make the encoded features of simulated and real data indistinguishable \citep{tzeng2017adversarial}.
This step allows us to learn $P(\vx_r|\vx_e, \vy_r)$ in a space with reduced domain discrepancy during the second stage.

Once trained, we pair $P(\vx_r|\vx_e, \vy_r)$ with the real data's encoder-decoder to generate real-style point clouds grounding on real ego-car's data, so that \emph{we can develop collaborative perception algorithms without real collaborative data.}
More specifically, given a real ego-car's perception $\vx_e$ and label $\vy_e$, we first translate them to center around the reference position, followed by optionally injecting object labels into $\vy_e$ to make it realistic from the reference viewpoint.

We extensively validate \ours on multiple datasets, all in an \emph{offline} setting. Empirical results demonstrate \ours's effectiveness in generating high-quality reference data to aid in the development of collaborative perception. In particular, we show that a conditioned diffusion model trained solely on simulated data (\eg, OPV2V~\citep{xu2022opv2v}) can already turn a real single-agent dataset into a real-alike collaborative one. As such, one can train collaborative perception algorithms for real test data without real training data. We further generate the ``ColWaymo'' dataset by training \ours on real single-agent Waymo data~\citep{sun2020waymo} and simulated OPV2V data~\citep{xu2022opv2v}, thus turning the former into a collaborative one. Collaborative perception backbones pre-trained on the large-scale, semi-synthetic ColWaymo data demonstrate a remarkable transferability. They notably improve collaborative perception developed in the downstream tasks (\eg, V2V4Real~\citep{xu2023v2v4real}), even in a many-shot setting.  

In sum, our key contributions are:
\begin{itemize}
\item We propose a new research direction to aid in the development of collaborative perception, \emph{generating real-looking sensory data from potentially any viewpoint in a real driving scene}, capable of turning a single-agent dataset into a collaborative one. This has the potential to scale up collaborative autonomous driving (CAV).
\item We present \ours, a two-stage training recipe with domain adaptation in between to learn the generative model using simulated collaborative data and real single-agent data.
\item Extensive experiments demonstrate \ours's effectiveness in aiding in collaborative perception across various scenarios.
\end{itemize}

\section{Related Work}
\label{sec:related}
\textbf{Collaborative autonomous driving (CAV)} offers significant benefits, including extended perception range by sharing sensor information to detect objects beyond a single vehicle's field of view.
Despite these benefits, collecting large-scale, real-world datasets for CAV poses significant challenges due to the complexity and cost of deploying multiple instrumented vehicles in diverse environments. Existing datasets often have limitations that restrict the scope of CAV research. For instance, OPV2V~\cite{xu2022opv2v} and V2X-Sim~\cite{li2022v2xsim} rely on simulations using the CARLA simulator~\citep{Dosovitskiy17carla}, allowing controlled scenarios with multiple vehicle agents, yet lacking real-world variability. V2V4Real~\cite{xu2023v2v4real}, on the other hand, provides real-world data, yet it only contains two collaborating agents, which restricts the exploration of more complex multi-agent interactions. Similarly, the DAIR-V2X~\cite{dair-v2x} dataset also offers real-world data including vehicle-to-infrastructure (V2I) and V2V, but it mainly focuses on V2I scenarios and still has a limited number of collaborating vehicles. To overcome current limitations, we propose a new research direction---generating realistic point clouds given the ego's perspective. 

\nbf{3D generation} Various works have explored 3D scene generation, such as LiDARGen~\cite{zyrianov2022learning-lidargen}, R2DM~\cite{nakashima2024lidar-r2dm}, LiDM~\cite{ran2024towards}, UltraLiDAR~\citep{xiong2023learning}, and RangeLDM~\cite{hu2025rangeldm}. Additionally, LidarDM~\cite{zyrianov2024lidardm} and Text2LiDAR~\cite{wu2024text2lidar} investigate conditional scene generation, with the former relying on hand-crafted map layouts and the latter conditioning on text inputs. However, all these methods focus on ego-centric generation. We propose a new research direction for CAV, aiming to generate realistic and consistent scenes conditioned on the ego agent's real point clouds---an area that remains largely unexplored.

\nbf{Diffusion models}
Diffusion models have recently advanced generative modeling for high-quality LiDAR point clouds and images. Denoising Diffusion Probabilistic Models (DDPMs)~\cite{ho2020denoising} outperform traditional Generative Adversarial Networks (GANs)~\cite{goodfellow2014generative} and further enhance efficiency for LiDAR generation. In autonomous driving applications, methods like RangeLDM~\cite{hu2025rangeldm}, LidarDM~\cite{zyrianov2024lidardm}, and LiDARGen~\cite{zyrianov2022learning-lidargen} apply diffusion models for realistic LiDAR scene generation. In this paper, we utilize its generation capability to study our proposed problem.

\begin{figure}
\centering
\includegraphics[width=1.0\linewidth]{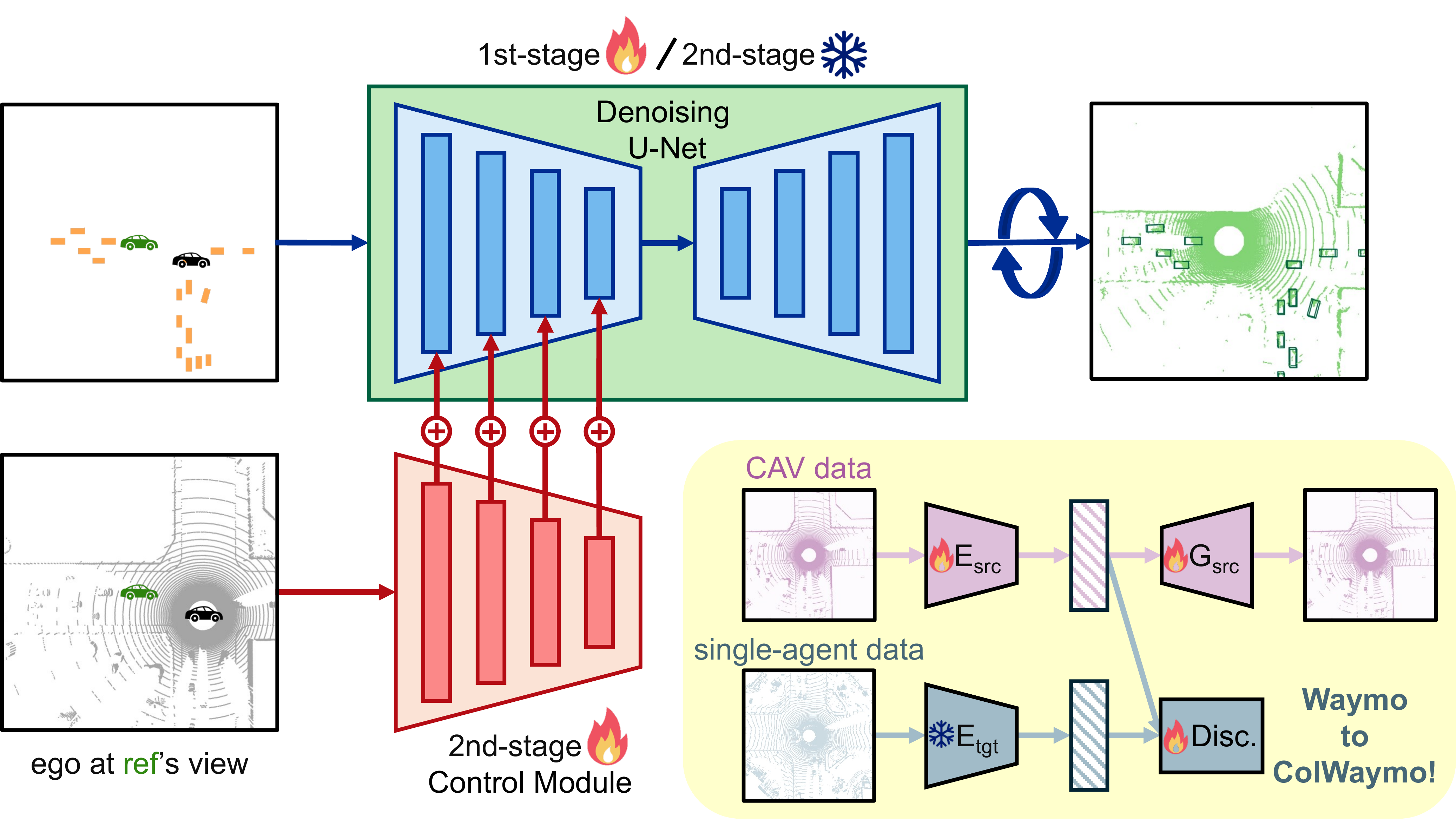}
\vspace{-6mm}
\caption{
\label{fig:pipeline}
\small \textbf{Illustration of \ours's conditioned generative model and training process.} We propose a two-stage training procedure. The first stage maximizes the generation capability by conditioning solely on object locations (using real single-agent target data), while the second stage grounds the generation on the ego-car's perspective to match semantics and layouts (using simulated CAV data). Additionally, we introduce a discriminator to adapt simulated CAV features to the real target domain, 
making the trained model readily applicable to the target domain after the second stage. 
}
\vspace{-5mm}
\end{figure}

\section{Transfer Your Perspective (\ours)}

\label{sec:method}
In this paper, we introduce a new research direction to advance collaborative autonomous driving (CAV): \emph{generating LiDAR point clouds from different perspectives within the same scene as the ego agent}, aiming to reduce the tedious efforts of collecting data for CAV. We begin by defining the proposed problem in \cref{sec:setup}. In \cref{sec:input}, we discuss the representations of the inputs, including point clouds and semantic information. \cref{sec:typ} outlines the pipeline developed to address this problem. Finally, in \cref{sec:adapt}, we demonstrate how this capability can be applied to datasets that only have labeled ego agents, \eg, Waymo Open Dataset (WOD)~\citep{sun2020waymo}.

\subsection{Problem Setup}
\label{sec:setup}
Given the perception data $\vx_e$ from an \textbf{e}go agent, we aim to build a model $P(\vx_r|\vx_e)$ to generate new perception data $\vx_r$ seen from a different \textbf{r}eference location and perspective within a communication range. Here, $\vx$ represents a LiDAR point cloud. This task presents several challenges, including potential information gaps between the two views, the need for alignment within the commonly perceivable area, and ensuring realism in the generated data, as discussed in \cref{sec:intro}.

To address these challenges, we extend the original problem by incorporating semantic information, such as object bounding boxes, represented as $P(\vx_r|\vx_e, \vy_r)$. We assume the availability of this information around the ego agent, denoted by $\vy_e$. For example, most of the existing single-agent datasets provide object labels. 
This semantic information can be easily translated and edited to become $\vy_r$ by tools like traffic re-players~\citep{gulino2024waymax}, making it a key to bridge the information gap between the ego and reference views, enabling the reference agent to ``see'' those objects and surfaces beyond the ego’s view---a core concept in CAV. %
Additionally, by aligning generated point clouds with object locations, this extension facilitates our goal of scaling up CAV development.
In essence, the generated $(\vx_r, \vy_r)$ pairs can be used flexibly alongside $(\vx_e, \vy_e)$ in various CAV applications, like directly using them to train collaborative perception algorithms.
 
In the following sections, we describe how $\vx$ and $\vy$ are encoded, followed by our approach to tackling the problem.

\subsection{Representations and Embeddings}
\label{sec:input}

\nbf{LiDAR} There are multiple ways to represent point clouds in continuous 3D space, such as coordinates (\ie, x, y, and z)~\citep{qi2017pointnet,qi2017pointnet++,shi2019pointrcnn}, range images~\citep{qi2017pointnet, wu2018squeezeseg, milioto2019rangenet++}, and voxelization~\citep{maturana2015voxnet,lang2019pointpillars,yan2018second,zhou2018voxelnet}. To better align with spatial control from object locations, we follow \citep{xiong2023ultralidar} to voxelize point clouds using a pre-defined grid and record voxel occupancy. \citep{xiong2023ultralidar} also highlights that this representation can naturally handle varying point densities and only minimally impacts LiDAR generation, with some precision trade-offs during voxelization. In short, we represent a point cloud by $\vx \in \sR^{H \times W \times C}$. 

To avoid the computational cost of 3D convolutions, we convert $\vx$ into Bird’s-Eye-View (BEV) images by treating the height dimension (\ie, $C$) as feature channels for 2D convolutions. This approach has been widely adopted in self-driving perception \citep{chen2017multi,zhang2018efficient,xiong2023learning}, allowing the use of 2D image-based model architectures and algorithms.

\nbf{Semantic information} Given that the point clouds are represented as BEV images, it is also intuitive to represent object locations in BEV. We create binary object maps by considering only the x and y coordinates of 3D bounding boxes, resulting in $\vy \in \sR^{H \times W \times 1}$. (We could further extend it by including category information.)

\nbf{Feature embedding} Following the literature~\citep{rombach2022high}, we encode input tensors using a Vector Quantized-Variational Autoencoder (VQ-VAE)~\citep{van2017neural}, comprising an encoder $E$, a quantization function $Q$ for feature vectors on spatial grids, and a decoder $G$. Formally, we obtain $\vx^f=E_x(\vx) \in \sR^{h \times w \times c_x}$ by the encoder, map each $c_x$-dimensional vector to a learnable code to obtain $\vx^z=Q_x(\vx^f) \in \sR^{h \times w \times c_x}$, and generate outputs using the decoder $\hat{\vx}=G_x(\vx^z) \in \sR^{H \times W \times C}$, where $h$ and $w$ are spatial resolutions of the feature map down-sampled from $H$ and $W$, and $c_x$ is the number of channels. 

The VQ-VAE model is trained end-to-end by minimizing:
\begin{equation}
    \Ls_{\text{vq}} = \Ls_{\text{rec}} + \lVert\text{sg}[E_x(\vx)] - \vx^z\rVert^2_2 + \lVert\text{sg}[\vx^z] - E_x(\vx)\rVert^2_2,
\label{eq:vqvae}
\end{equation}
where {sg[$\cdot$]} and $\Ls_{\text{rec}}$ denote the stop-gradient operation and the reconstruction loss, respectively.
As our point cloud representation is binary occupancy $\vx \in \{0, 1\}^{H\times W\times C}$, binary cross-entropy is a natural choice for $\Ls_{\text{rec}}$. However, due to the sparsity of point clouds, this results in an imbalanced loss. To address this, we adopt the Focal Loss (FL)~\citep{ross2017focal}:
\begin{equation}
    \ell_{\text{FL}}(\vx_i, \hat{\vx_i})=\left\{
    \begin{array}{ll}
        -(1-\hat{\vx_i})^{\gamma}\log(\hat{\vx_i}) & \text{if } \vx_i=1 \\
        -\hat{\vx_i}^{\gamma}\log(1-\hat{\vx_i}) & \text{otherwise}, \\
    \end{array} 
    \right.
\label{eq:focal}
\end{equation}
\begin{equation}
    \Ls_{\text{rec}} =\sum_{i=0}^{M}\ell_{\text{FL}}(\vx_i, \hat{\vx_i}),
\label{eq:rec}
\end{equation}
where $i$ is the voxel index and $M$ is the number of voxels.

For object locations $\vy$, we use the same approach to train a separate VQ-VAE model $E_y$, $Q_y$, and $G_y$. In the subsequent subsections, we use $\vx^f \in \sR^{h \times w \times c_x}$ and $\vy^f \in \sR^{h \times w \times c_y}$ as feature embeddings for learning a latent diffusion model.

\subsection{Transfer Your Perspective by Generation}
\label{sec:typ}

We model the distribution $P(\vx_r|\vx_e, \vy_r)$ defined in \cref{sec:setup} by a conditioned generative model and \emph{train it in two stages}. We assume access to $(\vx_r, \vx_e, \vy_r)$ tuples as training data. We detail the training data preparation in \cref{sec:adapt} and \cref{sec:exp}.
 
In the first stage, we aim to maximize the generation capability by providing minimal conditions, \ie, only bounding boxes, without further constraints on the scene. This encourages the model to produce $P(\vx|\vy)$ with high flexibility. Second, building on this model, we incorporate the ego agent's point cloud as an additional cue and ground the generation process on it to ensure semantic and layout consistency between the generated and ego agent's point clouds. 
Put together, this two-stage training procedure enables the learned $P(\vx_r|\vx_e, \vy_r)$ model to generate perception data at areas beyond the commonly visible regions between two views. 
The model and training process is illustrated in \cref{fig:pipeline}.

More importantly, this approach enables us to leverage existing labeled single-agent datasets such as KITTI~\citep{geiger2012kitti}, NuScenes~\cite{caesar2020nuscenes}, and Waymo Open Dataset (WOD)~\citep{sun2020waymo}, since the first-stage training only requires ego-centric LiDAR point clouds and their corresponding semantic information (see \cref{sec:adapt}). The $(\vx_e, \vx_r)$ data pairs seen from the ego and the reference agents are needed only in the second stage.

In the following, we elaborate on each training stage.

\nbf{Stage 1: Generation with semantic information}
The goal of this stage is to equip the model with strong generation capabilities of point clouds given spatial conditions, preparing it for the next stage. As discussed in \cref{sec:input}, the point clouds and object locations are embedded by VQ-VAEs~\citep{van2017neural}, denoted as $\vx^f=E_x(\vx) \in \sR^{h \times w \times c_x}$ and $\vy^f=E_x(\vy) \in \sR^{h \times w \times c_y}$. The image-like feature maps allow us to adopt existing generation algorithms for 2D images~\citep{dhariwal2021diffusion,ho2020denoising,kingma2021variational}. In this paper, we apply one of the most popular generative models, the Latent Diffusion Model (LDM)~\citep{rombach2022high}, for conditioned generation $P(\vx|\vy)$. LDM seeks to model a data distribution by iteratively denoising variables that are initially sampled from a Gaussian. The objective is:
\begin{equation}
    \Ls_{\text{LDM}} = \sE_{\vx^f, \vy^f, \epsilon \sim \gN(0,1), t}\left[\left\|\epsilon-\epsilon_\theta\left(\vx^f_t, \vy^f, t\right)\right\|_2^2\right],
\label{eq:ldm}
\end{equation}
where, $\epsilon_\theta$ is a UNet~\citep{ronneberger2015u} backbone and $t$ is the timestamp.

\nbf{Stage 2: Generation for new perspectives} In this stage, we aim to ground the generation of point clouds by incorporating the ego agent's perception. To retain the generation capability from stage 1, we freeze the learned $P(\vx|\vy)$ 
and introduce a learnable lightweight control module to inject additional cues, following T2I-Adapter~\citep{mou2024t2i}, as shown in \cref{fig:pipeline}. 

Formally, let $\vx_{e'}$ denote 
the \emph{translated} and \emph{rotated} ego-agent's point cloud centered around the reference location and aligned with the reference orientation, as shown in \cref{fig:pipeline}; $\vx_{e'}^f=E_x(\vx_{e'}) \in \sR^{h \times w \times c_x}$ is the corresponding embedding. 
Let $F_{AD}$ denote the learnable control module, which takes $\vx_{e'}^f$ as input and outputs 
\begin{equation}
    \rmF_c = F_{AD}(\vx_{e'}^f),\label{eq:T2I-adapter}
\end{equation}
where $\rmF_{c}=\{\rmF_c^1, \rmF_c^2, \rmF_c^3, \rmF_c^4\}$ matches the size of the  
multi-scale features $\rmF_{enc}=\{\rmF_{enc}^1, \rmF_{enc}^2, \rmF_{enc}^3, \rmF_{enc}^4\}$ extracted from the frozen encoder of the UNet 
$\epsilon_\theta$. With these ingredients, T2I-Adapter influences the generation process by 
\begin{equation}
    {\rmF}_{enc}^{i} \leftarrow \rmF_{enc}^{i} + \rmF_{c}^{i},\ i\in \{1,2,3,4\},
\label{eq:t2i}
\end{equation}
which injects $\vx_{e'}^f$ embedding into each of the denoising steps. The second-stage objective for learning $F_{AD}$ is:
\begin{equation}
    \resizebox{0.91\hsize}{!}{$\Ls_{\text{LDM}} = \sE_{\vx_r^f, \vy_r^f, \vx_{e'}^f, \epsilon \sim \gN(0,1), t}\left[\left\|\epsilon-\epsilon_\theta\left(\vx^f_{r, t}, \vy^f, \vx_{e'}^f, t\right)\right\|_2^2\right]$},
\label{eq:ldm2}
\end{equation}
where the encoder of $\epsilon_\theta$ is modified as defined in \cref{eq:t2i}.

\subsection{From Single Agent to Collaborative Datasets}
\label{sec:adapt}
There are many publicly accessible large-scale real-world autonomous driving datasets for ego-centric 3D perception, like WOD~\citep{sun2020waymo}, NuScenes~\citep{caesar2020nuscenes}, KITTI-360~\citep{liao2022kitti360}, and PandaSet \citep{xiao2021pandaset}, whereas much fewer datasets exist for CAV due to the challenges in data collection as discussed in \cref{sec:intro}.

This limitation inspired a bold idea: \emph{Can we transfer the generation capability learned from paired CAV data to these existing labeled single-agent datasets?} Achieving this would scale up CAV datasets substantially, offering immense benefits to the research community. However, this goal is challenging due to domain gaps between datasets, such as variations in LiDAR sensors, point cloud densities, data patterns, and data collection environments. Below, we describe our approach to addressing these challenges.

\nbf{Problem setup} Our target domain is a single-agent dataset, while the source domain consists of paired perception and semantic information from ego and reference perspectives, denoted as $(\vx_r, \vx_e, \vy_r)$. We demonstrate that we can transfer knowledge from a simulated CAV dataset (e.g., OPV2V~\cite{xu2022opv2v}) to a real-world ego-centric dataset (e.g., WOD~\citep{sun2020waymo}).

\nbf{Two-Stage training is the key} As described in \cref{sec:typ}, we decompose the generative training into two stages: first learning $P(\vx|\vy)$, followed by $P(\vx_r|\vx_e, \vy_r)$. The second stage keeps $P(\vx|\vy)$ frozen by adding an adapter. This strategy enables us to use target domain data in the first stage. The resulting $P(\vx|\vy)$ would generate target-like point clouds.

\nbf{Minimizing domain gaps} 
In the first stage, we have learned VQ-VAE encoder-decoders and $P(\vx|\vy)$ in the target domain (single-agent). 
In the second stage, we bring in CAV data to guide the generation on how to condition on ego-agent's perception. 
To minimize the gap of $\vx^f$ between two domains---\emph{$\vx^f$ is what the generative model is optimized to generate}---we adopt a GAN-style discriminator~\citep{lim2017geometric,goodfellow2014generative,ganin2016domain,tzeng2017adversarial}. 
Specifically, we aim to make feature embeddings extracted from the source data and the target data indistinguishable; namely, to confuse a discriminator $D$ trained to differentiate them.
We freeze all parameters of the target domain's encoder $E_{tgt}$, codebook $Q_{tgt}$, and decoder $G_{tgt}$ in the VQ-VAE. For the source domain, we initialize its $E_{src}$ and $G_{src}$ by $E_{tgt}$ and $G_{tgt}$ while \emph{reusing the frozen codebook $Q_{tgt}$}. We then learn the discriminator $D$ and the source domain's encoder $E_{src}$ and decoder $G_{src}$ in an interleaving fashion by minimizing the following objective functions: 
\begin{equation}
\begin{split}  \text{For $D$: \hspace{2pt}}\Ls_D=\hspace{11pt}&\sE_{\vx \sim P_{tgt}}\left[\text{max}(0, 1 - D(E_{tgt}(\vx)))\right] \\
    + &\sE_{\vx \sim P_{src}}\left[\text{max}(0, 1 
 + D(E_{src}(\vx)))\right];
\end{split}
\label{eq:hinge}
\end{equation}
\begin{align}
\text{For VQ-VAE:\hspace{5pt}} & \Ls_{\text{vq}} = \Ls_{\text{rec}} + \Ls_{src} + \lVert\text{sg}[\vx^z] - E_{src}(\vx)\rVert^2_2,
\label{eq:vqvae2}\\
& \text{where\hspace{5pt}} \Ls_{src} = - \sE_{\vx \sim P_{src}}D(E_{src}(\vx)).
\end{align}
The other terms in \cref{eq:vqvae2} follow those in \cref{eq:vqvae}, where for $\vx \sim P_{src}$, $\hat{\vx}=G_{src} \circ Q_{tgt} \circ E_{src}(\vx)$ in $\Ls_{rec}$ . 
We note that $\Ls_D$ takes in two streams of data (\ie, source and target) and is defined by the Hinge Loss~\citep{lim2017geometric} on $\vx^f$; $\Ls_{\text{vq}}$ takes in only the source CAV data. We also note that embeddings for object locations does not require adaptation. 

After this adaptation step, we proceed into the second stage of generative training, where we drop the discriminator and train the control module defined in \cref{eq:T2I-adapter} by minimizing \cref{eq:ldm2}. The feature embeddings are produced by $E_{src}$ on top of $(\vx_r, \vx_e, \vy_r) \sim P_{src}$.

\nbf{Enhancement in target domains} With the discriminator and adversarial training,
we reduce the gap between source and target domain data, successfully adapting \ours to the single-agent dataset, as shown in \cref{exp:colwaymo}. To further improve the conditioned generation quality in the target domain, a further fine-tuning stage in it is desired. However, \emph{how can we do so without having reference agents' point clouds in the target domain?} Here, we propose two simple solutions.

Recall that in training $P(\vx_r|\vx_e, \vy_r)$ using the source CAV data, point clouds show up in both the input as conditions and the outputs as supervisions. In other words, the domain gap exists on both sides, and we aim to reduce it by further fine-tuning \emph{using the target single-agent dataset}.
First, to enhance the output generation in the target domain, we set ego's point clouds to empty, \ie, $\vx_e={0}^{H \times W \times C}$ for $(\vx_r, \vy_r) \sim P_{tgt}$. Then, to enable adaptation on inputs, we apply self-training~\citep{pan2023towards,zou2018unsupervised,zoph2020rethinking,lee2013pseudo}, randomly sampling a reference location in the scene to generate pseudo-point clouds for fine-tuning, \ie, $(\hat{\vx_r}, \vx_e, \vy_r)$ for $(\vx_e, \vy_r) \sim P_{tgt}$. We use the above target domain data and the source domain data to jointly fine-tune the control module (\cf \cref{sec:typ}, \cref{eq:t2i}, and \cref{eq:ldm2}) further. Empirical results in~\cref{exp:colwaymo} show the effectiveness of the proposed solution in turning single-agent datasets into collaborative datasets for CAV development.

\section{Experiments}
\label{sec:exp}
\subsection{Setup}
\nbf{Datasets}
In this study, we primarily use three datasets: OPV2V~\citep{xu2022opv2v}, V2V4Real~\citep{xu2023v2v4real}, and WOD~\citep{sun2020waymo}. OPV2V is a simulation-based CAV dataset that includes over 70 diverse driving scenes and more than 11,000 frames. We utilize this dataset initially to validate our proposed research problem and later to guide the generation training on single-agent datasets. V2V4Real is a real-world CAV dataset with approximately 20,000 LiDAR frames, which we use to validate \ours in a real-world CAV context, treating it as a single-agent dataset (i.e., without using any labels to train \ours). Lastly, we adopt WOD to scale up \ours, one of the largest real-world single-agent datasets available.

\begin{table}[t]
\centering
\resizebox{.9\linewidth}{!}{
\setlength{\tabcolsep}{6px}
\begin{tabular}{llcccc}
\toprule
\multirow{2}{*}{Method} & \multirow{2}{*}{Train Data} & \multicolumn{4}{c}{AP 0.5} \\ \cmidrule(lr){3-6} 
& & s & m & l & all \\ \midrule

No Fusion & ego's gt only & 0.67 & 0.41 & 0.13 & 0.40 \\
\cdashline{1-6} \noalign{\vskip2pt}

& gt~~~~~~(\textbf{oracle}) & 0.76 & 0.42 & 0.31 & 0.49 \\
 \rowcolor{lightcyan}
\cellcolor{white}\multirow{-3}{*}{Early Fusion~\citep{chen2019cooper}} & \ours (\textbf{ours}) & 0.75 & 0.38 & 0.29 & 0.46 \\ 

 \cdashline{1-6} \noalign{\vskip2pt}
& gt~~~~~~(\textbf{oracle}) & 0.74 & 0.57 & 0.36 & 0.55 \\
 \rowcolor{lightcyan}
\cellcolor{white}\multirow{-3}{*}{Late Fusion~\citep{xu2022opv2v}} & \ours (\textbf{ours}) & 0.71 & 0.49 & 0.32 & 0.50 \\ 

\cdashline{1-6} \noalign{\vskip2pt}
& gt~~~~~~(\textbf{oracle}) & 0.94 & 0.80 & 0.62 & 0.77 \\
 \rowcolor{lightcyan}
\cellcolor{white}\multirow{-3}{*}{AttFuse~\citep{xu2022opv2v}} & \ours (\textbf{ours}) & 0.90 & 0.73 & 0.56 & 0.72 \\

\cdashline{1-6} \noalign{\vskip2pt}
& gt~~~~~~(\textbf{oracle}) & 0.87 & 0.71 & 0.50 & 0.71 \\
 \rowcolor{lightcyan}
\cellcolor{white}\multirow{-3}{*}{V2X-ViT~\citep{xu2022v2xvit}} & \ours (\textbf{ours}) & 0.84 & 0.65 & 0.40 & 0.65 \\ \hline
\bottomrule
\end{tabular}
}
\centering
\captionsetup{width=1.\linewidth}
\vspace{-2mm}
\caption{
\label{tab:opv2v}
\small \textbf{Results on OPV2V.} We compare the performance using reference generated by \ours versus ground-truth (oracle). The comparable results demonstrate the quality of our generated data.}
\vspace{-3mm}
\end{table}

\begin{table}
\centering
\resizebox{.9\linewidth}{!}{
\setlength{\tabcolsep}{6px}
\begin{tabular}{llcccccccccccccccc}
\toprule
\multirow{2}{*}{Method} & \multirow{2}{*}{Train Data} & \multicolumn{4}{c}{{AP 0.5}} \\ \cmidrule(lr){3-6} 
& & s & m & l & all \\ \midrule

No Fusion & ego's gt only & 0.71 & 0.33 & 0.08 & 0.46 \\
\cdashline{1-6} \noalign{\vskip2pt}

& gt~~~~~~(\textbf{oracle}) & 0.79 & 0.42 & 0.45 & 0.60 \\
 \rowcolor{lightcyan}
\cellcolor{white}\multirow{-3}{*}{Early Fusion~\citep{chen2019cooper}} & \ours (\textbf{ours}) & 0.76 & 0.36 & 0.30 & 0.53 \\ 

 \cdashline{1-6} \noalign{\vskip2pt}
& gt~~~~~~(\textbf{oracle}) & 0.79 & 0.53 & 0.56 & 0.67  \\
 \rowcolor{lightcyan}
\cellcolor{white}\multirow{-3}{*}{Late Fusion~\citep{xu2022opv2v}} & \ours (\textbf{ours}) & 0.77 & 0.48 & 0.51 & 0.63 \\ 

 \cdashline{1-6} \noalign{\vskip2pt}
& gt~~~~~~(\textbf{oracle}) & 0.80 & 0.45 & 0.36 & 0.61  \\
 \rowcolor{lightcyan}
\cellcolor{white}\multirow{-3}{*}{AttFuse~\citep{xu2022opv2v}} & \ours (\textbf{ours}) & 0.75 & 0.45 & 0.26 & 0.56 \\ 

\cdashline{1-6} \noalign{\vskip2pt}
& gt~~~~~~(\textbf{oracle}) & 0.81 & 0.49 & 0.30 & 0.61   \\
 \rowcolor{lightcyan}
\cellcolor{white}\multirow{-3}{*}{V2X-ViT~\citep{xu2022v2xvit}} & \ours (\textbf{ours}) & 0.76  & 0.38  & 0.17  & 0.48   \\ \hline
\bottomrule
\end{tabular}
}
\centering
\captionsetup{width=1.\linewidth}
\vspace{-2mm}
\caption{
\label{tab:real}
\small \textbf{Results on V2V4Real.} We compare the performance using reference generated by \ours versus ground-truth (oracle). Note that \emph{we do not use V2V4Real data to train the generation}, treating it as a single-agent dataset. The comparable results demonstrate the quality of our generated data and the transferability of \ours.}
\vspace{-5mm}
\end{table}

\begin{figure}
\centering
\includegraphics[trim={0cm, 0cm, 0cm, 0cm},clip,width=.95\linewidth]{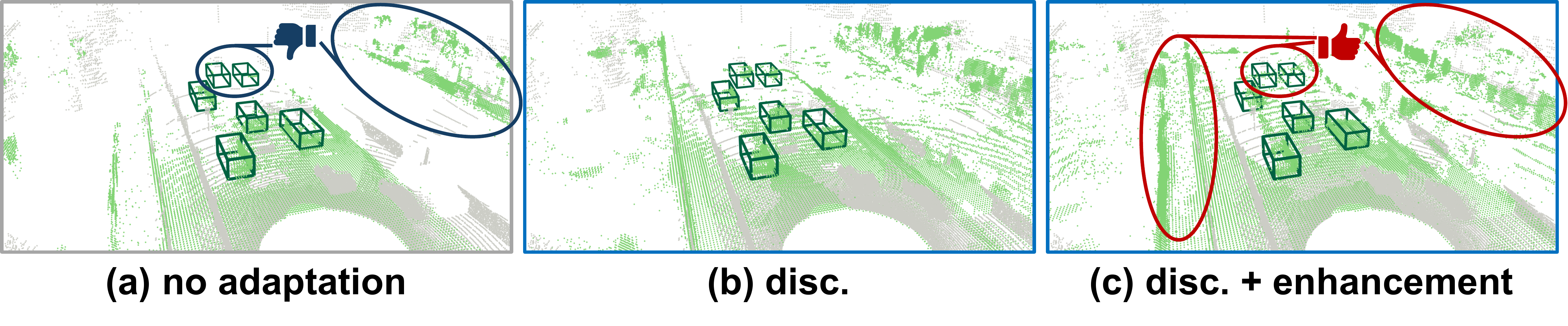}
\vspace{-4mm}
\caption{
\label{fig:adapt}
\small \textbf{Qualitative results on enhancement in the target domain.} Generated point cloud (\textcolor{customgreen}{green}) has better quality with the enhancement given ego (\textcolor{gray}{gray}) from Waymo (\cf \cref{sec:adapt}).}
\vspace{-5mm}
\end{figure}

\begin{table*}
\centering
\resizebox{.9\linewidth}{!}{
\setlength{\tabcolsep}{4px}
\begin{tabular}{llcccccccccccccccc}
\toprule
\multirow{2}{*}{Method} & \multirow{2}{*}{Pre-Train} & \multicolumn{4}{c}{FT. on 0 Scene} & \multicolumn{4}{c}{FT. on 5 Scene} & \multicolumn{4}{c}{FT. on 10 Scene} & \multicolumn{4}{c}{FT. on 32 Scene} \\ \cmidrule(lr){3-6} \cmidrule(lr){7-10} \cmidrule(lr){11-14} \cmidrule(lr){15-18}
& & s & m & l & all & s & m & l & all & s & m & l & all & s & m & l & all \\ \midrule

No Fusion & ego's gt only &  &  &  &  & 0.44 & 0.16 & 0.06 & 0.29 & 0.66 & 0.29 & 0.08 & 0.42 & 0.71 & 0.34 & 0.08 & 0.47 \\
\cdashline{1-18} \noalign{\vskip2pt}

& scratch & & & &  & 0.44 & 0.18 & 0.16 & 0.30 & 0.73 & 0.28 & 0.24 & 0.48 & 0.79 & 0.42 & 0.45 & 0.60 \\
& OPV2V & \textbf{0.54} & 0.20 & 0.07 & 0.31 & 0.32 & 0.18 & 0.15 & 0.25 & 0.73 & 0.32 & 0.31 & 0.50 & 0.80 & 0.47 & \textbf{0.54} & \textbf{0.65} \\
 \rowcolor{lightcyan}
\cellcolor{white}\multirow{-3}{*}{Early Fusion~\citep{chen2019cooper}} & ColWaymo (\textbf{ours}) & 0.50 & \textbf{0.24} & \textbf{0.24} & \textbf{0.35} & \textbf{0.68} & \textbf{0.34} & \textbf{0.32} & \textbf{0.51} & \textbf{0.78} & \textbf{0.39} & \textbf{0.33} & \textbf{0.57} & \textbf{0.83} & \textbf{0.48} & 0.50 & \textbf{0.65} \\ 

 \cdashline{1-18} \noalign{\vskip2pt}
& scratch &  &  & &  & 0.44 & 0.24 & 0.33 & 0.34 & 0.72 & 0.42 & \textbf{0.51} & 0.58 & 0.79 & 0.53 & 0.56 & 0.67 \\
& OPV2V & \textbf{0.60} & \textbf{0.27} & \textbf{0.28} & \textbf{0.44} & 0.55 & 0.26 & 0.38 & 0.42 & 0.73 & 0.42 & \textbf{0.51} & 0.59 & 0.78 & 0.55 & 0.58 & 0.67 \\
 \rowcolor{lightcyan}
\cellcolor{white}\multirow{-3}{*}{Late Fusion~\citep{xu2022opv2v}} & ColWaymo (\textbf{ours}) & 0.40 & 0.19 & 0.15 & 0.25 & \textbf{0.60} & \textbf{0.28} & \textbf{0.44} & \textbf{0.47} & \textbf{0.76} & \textbf{0.43} & \textbf{0.51} & \textbf{0.61} & \textbf{0.82} & \textbf{0.58} & \textbf{0.61} & \textbf{0.71} \\ 

\cdashline{1-18} \noalign{\vskip2pt}
& scratch &  &  &  &  & 0.40 & 0.15 & 0.10 & 0.28 & 0.70 & 0.30 & 0.17 & 0.47 & 0.80 & 0.45 & 0.36 & 0.61 \\
& OPV2V & 0.51 & 0.19 & 0.05 & 0.31 & 0.54 & 0.22 & 0.11 & 0.37 & 0.77 & 0.40 & 0.21 & 0.54 & 0.83 & 0.53 & 0.40 & 0.65 \\
 \rowcolor{lightcyan}
\cellcolor{white}\multirow{-3}{*}{AttFuse~\citep{xu2022opv2v}} & ColWaymo (\textbf{ours}) & \textbf{0.66} & \textbf{0.35} & \textbf{0.11} & \textbf{0.45} & \textbf{0.65} & \textbf{0.29} & \textbf{0.16} & \textbf{0.46} & \textbf{0.83} & \textbf{0.46} & \textbf{0.33} & \textbf{0.61} & \textbf{0.88} & \textbf{0.58} & \textbf{0.53} & \textbf{0.72} \\

\cdashline{1-18} \noalign{\vskip2pt}
& scratch &  &  &  &  & 0.43 & 0.15 & 0.12 & 0.31 & 0.70 & 0.28 & 0.17 & 0.43 & 0.81 & 0.49 & 0.30 & 0.61 \\
& OPV2V & 0.51 & 0.24 & 0.07 & 0.33 & 0.48 & 0.23 & 0.16 & 0.35 & 0.76 & 0.38 & 0.22 & 0.53 & 0.81 & 0.49 & 0.35 & 0.61 \\
 \rowcolor{lightcyan}
\cellcolor{white}\multirow{-3}{*}{V2X-ViT~\citep{xu2022v2xvit}} & ColWaymo (\textbf{ours}) & \textbf{0.60} & \textbf{0.28} & \textbf{0.10} & \textbf{0.34} & \textbf{0.66} & \textbf{0.28} & \textbf{0.22} & \textbf{0.46} & \textbf{0.79} & \textbf{0.48} & \textbf{0.26} & \textbf{0.58} & \textbf{0.84} & \textbf{0.57} & \textbf{0.44} & \textbf{0.67} \\ \hline
\bottomrule
\end{tabular}
}
\centering
\captionsetup{width=1.\linewidth}
\vspace{-2mm}
\caption{
\label{tab:ft}
\small \textbf{Results with pre-training on collaborative Waymo.} We scale up \ours with exiting large-scale single-agent dataset WOD~\citep{sun2020waymo}, creating its collaborative version ``ColWaymo''. These data are used for pre-training and subsequently fine-tuned on V2V4Real~\citep{xu2023v2v4real}.  Results show significant improvement over training from scratch, highlighting the potential to reduce data collection efforts, scale up, and accelerate CAV development. Comparisons to simulated OPV2V~\citep{xu2022opv2v} pre-training further demonstrate the realism of our generated point clouds.
}
\vspace{-5mm}
\end{table*}

\begin{table}
\centering
\resizebox{.95\linewidth}{!}{
\setlength{\tabcolsep}{4px}
\begin{tabular}{lccccc}
\toprule
Method & Disc. & ST. & Dum. & JSD ($\downarrow$) & MMD ($\downarrow$) \\ \midrule
No Adaptation & & & & $0.26$ & $4.61\times 10^{-4}$ \\
\cdashline{1-6} \noalign{\vskip2pt}
& \checkmark & & & $0.16$ & $1.10$ $\times 10^{-4}$ \\
& \checkmark & \checkmark & & $0.17$ & $1.30$ $\times 10^{-4}$ \\
\cellcolor{white}\multirow{-3}{*}{Adaptation} & \checkmark & \checkmark & \checkmark & $0.16$ & $1.17$ $\times 10^{-4}$ \\ \hline
\bottomrule
\end{tabular}
}
\centering
\captionsetup{width=1.\linewidth}
\vspace{-2mm}
\caption{
\label{tab:dist}
\small \textbf{Ablation of proposed adaption method}. Results with adaptation outperform no adaptation, demonstrating an improvement in generation quality. Abbreviations in title refer to discriminator, self-training, and dummy ego, respectively (\cf \cref{sec:adapt}).}
\vspace{-4mm}
\end{table}

\begin{table}
\centering
\resizebox{.95\linewidth}{!}{
\setlength{\tabcolsep}{4px}
\begin{tabular}{ccccc}
\toprule
Num. Steps & Epochs 1st & Epochs 2nd & JSD ($\downarrow$) & MMD ($\downarrow$) \\ \midrule
1 & -- & 280 & $0.11$ & $6.45\times 10^{-5}$ \\
2 & 200 & 80 & $\bm{0.10}$ & $\bm{5.28 \times 10^{-5}}$ \\ \hline
\bottomrule
\end{tabular}
}
\centering
\captionsetup{width=1.\linewidth}
\vspace{-2mm}
\caption{
\label{tab:step}
\small \textbf{2-Stage training.} Results demonstrate the benefit of the proposed 2-stage training in generation quality on OPV2V~\citep{xu2022opv2v}. 
}
\vspace{-5mm}
\end{table}

\nbf{Evaluation metrics}
Following standard benchmarks~\citep{xu2022opv2v,xu2023v2v4real}, we report Average Precision (AP 0.5) across three distance ranges: 0-30 meters (short, denoted as s), 30-50 meters (medium, m), and 50-100 meters (long, l) and overall (all).

\nbf{Implementations}
For \ours training, we voxelize point clouds within the range $[-51.2, 51.2]$ meters in x and y dimensions and 4 meters in z (range is dataset-dependent) to create 3D volumes, $\vx \in \sR^{512 \times 512 \times 20}$. This is then encoded to $\vx^f \in \sR^{64 \times 64 \times 8}$ with VQ-VAE~\citep{van2017neural}. Similarly, semantic information $\vy \in \sR^{512 \times 512 \times 1}$ is encoded to $\vy^f \in \sR^{64 \times 64 \times 3}$. We train VQ-VAE for $\vx$ 120 epochs and for $\vy$ 10 epochs, with a batch size of 6 per GPU. Next, we train LDM with a batch size of 48 per GPU for 200 epochs in the first stage and 80 epochs in the second stage. During the second stage, we freeze the U-Net from the first stage and fine-tune only the control module, following~\citep{mou2024t2i}. For adaptation, we initialize the weights of encoder-decoder for CAV data (fake) with the model trained on single-agent data (real), freeze its codebook, and fine-tune it with a discriminator comprising 4 convolutional layers. The model processes an equal amount of fake and real per batch (i.e., 3 samples each). All training is conducted on 4 NVIDIA H100 GPUs. For experiments on benchmarks, we follow the convention to report results with Pointpillars~\citep{lang2019pointpillars}. We train / fine-tune all models with $8$ P100, with batch size $4$ per GPU. For V2X-ViT, we use $4$ H100 due to resources required.

\nbf{Inference of generation and post-processing}
We use DDIM scheduler~\citep{song2020denoising} for 200 steps. For post-processing, we apply modified Gumbel-Softmax~\citep{jang2016categorical} to the logits. Specifically, we define two thresholds, low and high. Scores below the low threshold are dropped, those above the high threshold are retained, and values in between are perturbed with Gaussian noise on logits. Hyperparameters are tuned on a small hold-out set. Finally, we convert the volume back to coordinates by using the centers of the occupied voxels.

\subsection{Do We Need Collaborative Perception?}
\label{sec:question}
To explore this question, we consistently report the performance without collaborative agents (denoted as ``ego’s gt only'' in \cref{tab:opv2v}, \cref{tab:real}, and \cref{tab:ft}). For example, in the V2V4Real dataset, a well-trained detector struggles with objects beyond 50 meters, achieving only $0.08$ AP at that range, though it performs well on nearby objects with an AP of $0.71$ (\cf first row of \cref{tab:ft}). simply combining the predictions from two under-trained detectors, we observe a notable boost in performance to $0.33$ (\cf ``scratch'' in late fusion, \cref{tab:real}), suggesting the significance of CAV.

\subsection{Point Clouds Generation on OPV2V}
\label{exp:opv2v}
\nbf{Setting}
To validate our research concept, we begin with the simulated dataset OPV2V~\citep{xu2022opv2v}. We split the original training set of $44$ scenes into two halves, using only the first $22$ scenes to train our generation model. For the remaining $22$ scenes, we generate point clouds for reference agents based on the ego agent’s point clouds and compare these results to ground-truth point clouds. We follow the benchmark to report performance on the validation set from Culver City.

\nbf{Baselines}
Following the literature~\citep{xu2022opv2v,xu2023v2v4real}, we adopt three common fusion algorithms for CAV: early, late, and intermediate fusion, which fuse input point clouds, extracted features, and predicted bounding boxes, respectively. For intermediate fusion, we adpot AttFuse~\citep{xu2022opv2v} and V2X-VIT~\citep{xu2022v2xvit}.

\nbf{Comparison to using ground-truth LiDAR}
In \cref{tab:opv2v}, we show that training CAV with point clouds generated by \ours achieves performance comparable to that obtained with ground-truth (oracle) across all methods (\eg, $0.46~\vs~0.49$ with early, $0.50~\vs~0.55$ with late, and $0.77~\vs~0.72$ with AttFuse~\citep{xu2022opv2v}, etc.) This result highlights the quality of the generated point clouds. In the \supp, we further demonstrate that with additional labeled data (from the first half of the split used for training generation), the gap to oracle performance can be further minimized.

\begin{figure*}[t]
\centering
\includegraphics[trim={0cm, 0cm, 0cm, 0cm},clip,width=.87\linewidth]{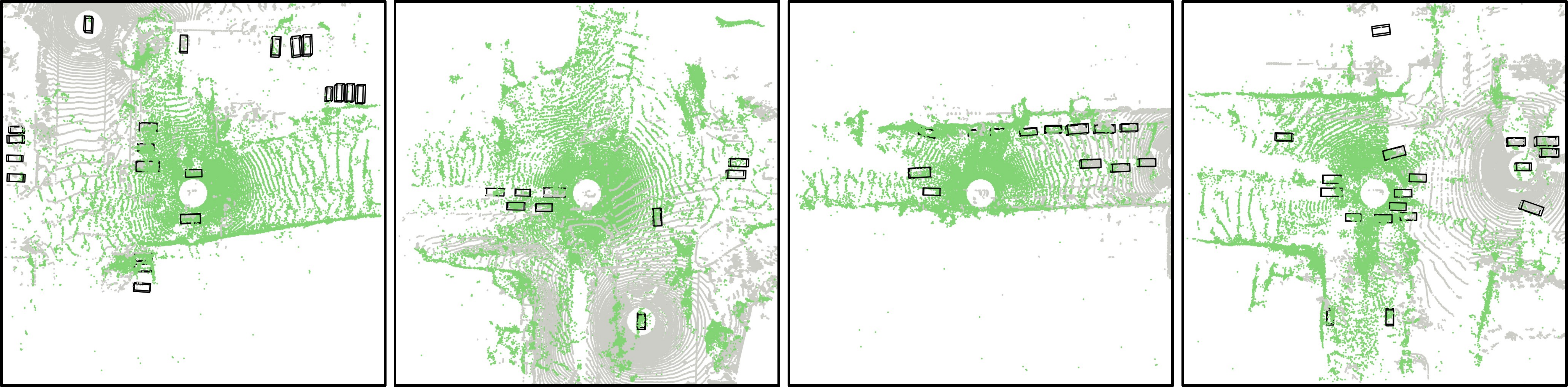}
\vspace{-3mm}
\caption{
\label{fig:colwaymo}
\small \textbf{Qualitative results on Collaborative Waymo.} The gray point clouds are from the original single-agent dataset and the green are generated by \ours conditioning on them.
}
\vspace{-3mm}
\end{figure*}

\subsection{Single-Agent to Collaborative Multi-Agent}
\label{exp:real}
\nbf{Setting} Building on our findings with simulated data (\cf \cref{exp:opv2v}), we validate our approach on the real-world dataset V2V4Real~\citep{xu2023v2v4real}, \emph{treating it as a single-agent dataset without access to reference agents’ point clouds}. Specifically, we use the model from \cref{exp:opv2v} to generate reference car's point clouds from the ego’s perspective and compare performance against ground-truth point clouds (oracle). Note that \emph{no V2V4Real data is used to train the generation}.

\nbf{Results on V2V4Real}
In \cref{tab:real}, we show the performance using generated reference point clouds is comparable to the oracle, supporting our idea of \emph{translating a single-agent dataset to multi-agent CAV}. Xu \etal previously explored OPV2V-to-V2V4Real adaptation, noting significant performance drops even with adaptation algorithms. Here, \ours offers a new solution to this challenge -- in a generative way, providing a fresh perspective on domain adaptation.
\begin{figure*}[ht]
\centering
\includegraphics[trim={0cm, 0cm, 0cm, 0cm},clip,width=.9\linewidth]{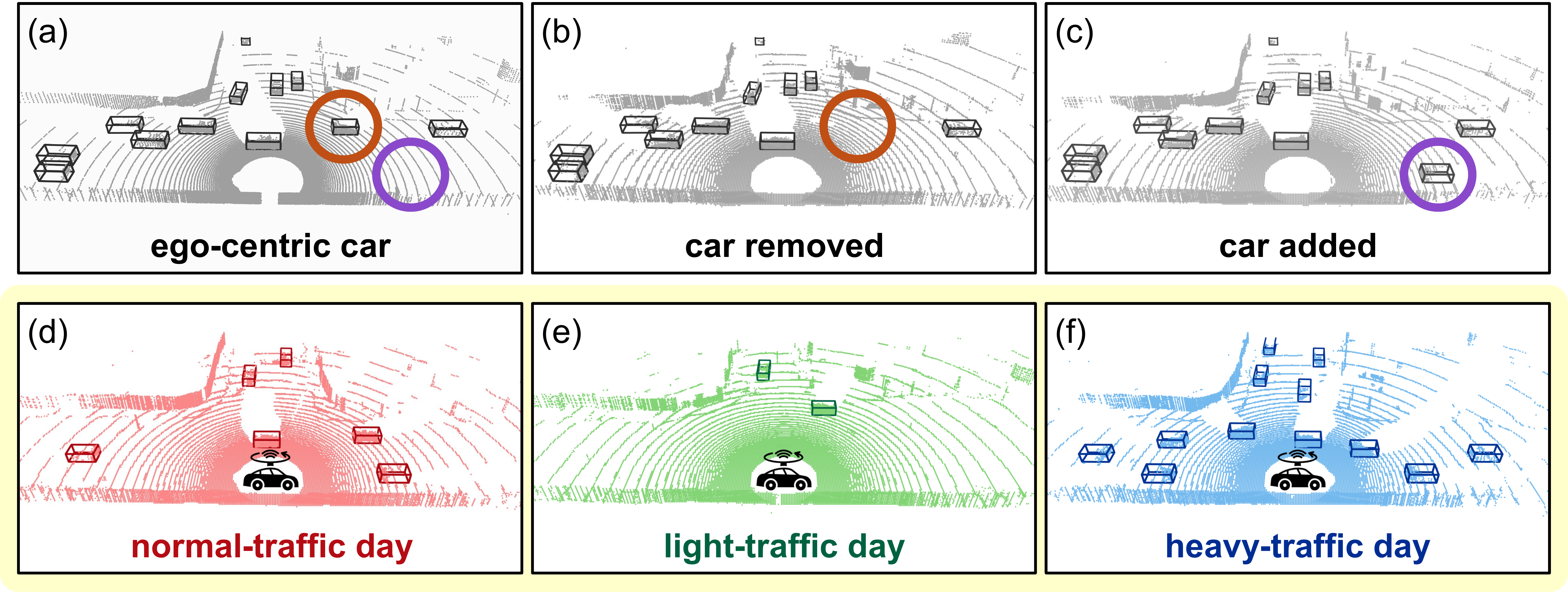}
\vspace{-3mm}
\caption{
\label{fig:qual}
\small \textbf{Qualitative results.} Our proposed \ours is capable of scene editing, by inputting the same point cloud but different object locations. We (b) remove and (c) add a car from (a) the original point cloud. Inspired by the idea of past traversals~\citep{you2022hindsight}, we apply completely different traffic conditions and generate (d), (e), and (f), to imagine driving through the same intersection.
}
\vspace{-5mm}
\end{figure*}

\subsection{ColWaymo: Scaling with Large-Scale Data}
\label{exp:colwaymo}

\nbf{Scaling CAV to the next level}
In \cref{exp:real}, we show the potential to create realistic reference's point clouds on real-world dataset V2V4Real~\citep{xu2023v2v4real}, through a simulation-guided training of \ours. Here, our goal is broader: to transform an existing large-scale single-agent dataset (\eg, WOD~\citep{sun2020waymo}) into a collaborative version. Using the pipeline described in \cref{sec:method} and illustrated in \cref{fig:pipeline}, we train the generative model on $17,400$ samples from the WOD training set. We then use its validation set to generate approximately $36,000$ collaborative samples by sampling reference locations from labeled vehicles. These generated samples serve to pre-train a model, which we subsequently fine-tune on V2V4Real~\citep{xu2023v2v4real}. For comparison, we also conduct pre-training on OPV2V~\citep{xu2022opv2v}.

\nbf{Results of fine-tuning on V2V4Real}
Following prior work~\citep{pan2023pre,xie2020pointcontrast,yin2022proposalcontrast,boulch2023also}, we perform pre-training followed by fine-tuning on limited labeled data to validate the potential of \ours for scaling up CAV development. As shown in \cref{tab:ft}, pre-training on our generated data yields significantly better performance than training from scratch. Additionally, compared to pre-training on the simulated dataset OPV2V, ColWaymo consistently achieves higher performance, highlighting the realism of the generated point clouds. These results demonstrate that \ours effectively transforms an ego-only dataset into a collaborative version, and potentially reducing the need for extensive CAV data collection.

\nbf{Generation quality by proposed adaptation} To assess the quality of generated data, we use two common metrics: Jensen-Shannon Divergence (JSD) and Maximum Mean Discrepancy (MMD). As shown in \cref{tab:dist}, our adaptation method improves JSD from $0.26$ to $0.16$ and MMD from $4.61$ to $1.17$ compared to the baseline without adaptation. Additionally, \cref{fig:adapt} visually illustrates this improvement.

\subsection{Qualitative Results}
We illustrate the proposed problem in \cref{fig:teaser}, which highlights the benefits of collaboration by allowing an agent to ``see'' beyond the limitations of its own sensors. \ours generates realistic point clouds from various viewpoints, enabling the agents to perceive previously occluded objects while preserving the overall scene semantics. \cref{fig:colwaymo} shows our ColWaymo, and \cref{fig:qual} further demonstrates the model’s capability for scene editing.

\section{Conclusion}
\label{sec:conclusion}
\vspace{-2.5mm}
We introduce a new research direction in collaborative driving and present the first solution. Empirical results show our approach can significantly reduce data collection and development efforts, advancing safer autonomous systems. 

\nbf{Limitations and future work}
Following existing benchmarks~\citep{xu2022opv2v, xu2023v2v4real}, \ours focuses on vehicle-like objects. Future work could extend it to broader objects and static entities (\eg, traffic signs, signals) essential for real-world traffic.

\section*{Acknowledgment}

This research is supported in part by grants from the National
Science Foundation (IIS-2107077, IIS-2107161). We are thankful for the generous support of the computational resources by the Ohio Supercomputer Center.

{\small
\bibliographystyle{ieeenat_fullname}
\bibliography{main}
}

\ifarxiv \clearpage \appendix In this \supp, we provide more details and experiment results in addition to the main paper:
\begin{itemize}
    \item \cref{sup:related}: provides more related works, including 3D scene generation, diffusion models, domain adaptation, and NeRF. 
    \item \cref{sup:datasets}: summarizes existing datasets for CAV.
    \item \cref{sup:opv2v}: conducts experiments on generated point clouds with additional ground-truths.
    \item \cref{sup:stat}: shows the statistical results of the experiments.
    \item \cref{sup:stage}: further demonstrates the generation quality of two-stage training with experiments on CAV setting.
    \item \cref{sup:qual}: shows more qualitative results. 
    \item \cref{sup:future}: concludes future directions.
\end{itemize}

\section{More Related Work}
\label{sup:related}

\nbf{3D Scene Generation}
As generative models have gained traction, recent research has increasingly focused on applying these methods to 3D point cloud data. Initially, the synthesis of point clouds was primarily limited to fixed-size data, such as single objects~\citep{achlioptas2018learning, fan2017point, shu20193d, valsesia2018learning}. However, recent advancements have extended beyond single-object generation to encompass entire 3D scenes. Early pioneering works in this domain employed generative adversarial networks (GANs)~\citep{goodfellow2014generative}, demonstrating the feasibility of 3D scene generation, albeit with significant challenges in quality and realism.

More recent efforts have aimed at improving the quality and realism of 3D scene generation. For instance, LiDARGen~\citep{zyrianov2022learning-lidargen} and UltraLiDAR~\citep{xiong2023ultralidar} leverage diffusion models to enhance scene quality, incorporating realistic effects like ray drop. However, these methods struggle to generate scenes based on user-defined conditions, such as specific locations or diverse traffic scenarios.

To address these limitations, works like LidarDM~\citep{zyrianov2024lidardm} have introduced more controllable scene generation using consecutive video frames and user-defined conditions. Similarly, Text2LiDAR~\citep{wu2024text2lidar} employs text prompts for conditioning, enabling diverse scene generation tailored to user inputs.

Other advancements prioritize flexibility, efficiency, and quality. R2DM~\citep{nakashima2024lidar-r2dm} proposes efficient training pipelines and a LiDAR completion framework that enhances scene quality. Meanwhile, RangeLDM~\citep{hu2025rangeldm} combines latent diffusion models with improved speed and quality for scene generation. LiDM~\citep{ran2024towards} synthesizes recent advancements to achieve state-of-the-art results in realistic 3D scene generation, balancing quality, realism, and user control with multiple conditioning inputs.

In this paper, we propose a novel research problem in the context of CAV: generating realistic point clouds for reference agents. This direction offers significant potential for the research community, addressing the critical challenge of data collection in CAV, which is inherently difficult and costly. Unlike existing works focused on ego-centric scene generation, our approach shifts the perspective to collaborative scenarios. We view this as a complementary research direction and are open to enhancing our proposed solution by adopting advancements in efficiency and controllability from ongoing work in 3D scene generation.

\nbf{Diffusion-based Generative Models}
Diffusion models (DMs)\citep{sohl2015deep} have made significant advancements in various domains, particularly in generating high-quality images. Initially, DMs were applied directly to raw pixel data, achieving remarkable results\citep{dhariwal2021diffusion, ho2020denoising, kingma2021variational}. To improve efficiency, Latent Diffusion Models (LDMs)~\citep{rombach2022high} operate in a compressed latent space, preserving visual quality while significantly reducing computational requirements. These approaches have found widespread application across diverse tasks, including 3D scene generation, as discussed in the previous section.

\nbf{Controllable Diffusion Models} Many existing works focus on controlling generative processes through text prompts, particularly in text-to-image (T2I) synthesis~\citep{nichol2022glide,ramesh2021zero,ding2021cogview,gafni2022make,rombach2022high,saharia2022photorealistic}. The predominant strategy involves performing denoising in feature space while integrating text conditions into the denoising process via a cross-attention mechanism. While these approaches achieve impressive synthesis quality, text prompts often lack reliable structural guidance for precise generation.

To address this limitation, several works improve structural control during generation. For instance, \citep{wang2022pretraining,hertzprompt,feng2023training,balaji2022ediff} explore methods to enhance structure guidance in text-driven synthesis. Meanwhile, works like \citep{mou2024t2i,li2025controlnet,zhang2023adding} introduce additional trainable modules built upon pre-trained T2I models to provide more targeted and controllable outputs.

In this paper, we leverage the approach proposed by \citep{mou2024t2i} during the second stage of our framework. This stage grounds the generation process, ensuring that the outputs align with given semantic cues.

\begin{table}
\resizebox{.99\linewidth}{!}{
\setlength{\tabcolsep}{4px}
\begin{tabular}{lcccccccc}
\toprule
\multirow{2}{*}{Datasets} & \multirow{2}{*}{Venue}   & \multirow{2}{*}{Real?} & \multicolumn{3}{c}{Agent} & \multirow{2}{*}{\# Cls} & \multirow{2}{*}{\# Frames} & \multirow{2}{*}{Mod.} \\ \cmidrule(lr){4-6} 
& & & dynamic & static & \# & & & \\ \midrule

OPV2V    & ICRA'22 & x     & o   & x   & 2-7       & 1            & 11.5k     & C, L     \\
V2X-Sim  & RA-L'22 & x     & o   & o   & 2-5       & 2            & 10k       & L        \\
V2XSet   & ECCV'22 & x     & o   & o   & 2-5       & 1            & 11.5k     & C, L     \\
DAIR-V2X & CVPR'22 & o     & x   & o   & 2         & 10           & 39k       & C, L     \\
V2V4Real & CVPR'23 & o     & o   & x   & 2         & 1 
  & 20k       & L        \\
MARS     & CVPR'24 & o     & o   & x   & 2         & x            & 15k       & C, L     \\ 
\cdashline{1-9} \rowcolor{lightcyan}
TYP's motivation & & semi-real & o & o & $\infty$ & 1 & 
 - & L \\ \bottomrule
\end{tabular}
}
\centering
\captionsetup{width=1.\linewidth}
\caption{
\label{tab:datasets}
\small \textbf{Existing datasets for CAV.} Real-world datasets are limited by the challenges of data collection. Our proposed research problem aims to address this issue.
}
\end{table}

\nbf{Domain Adaptation}
Unsupervised domain adaptation (UDA) has been extensively studied. A common approach for domain adaptation is to learn domain-invariant embeddings by minimizing the distributional differences between source and target domains~\citep{long2015learning,sun2016deep,tzeng2015simultaneous,tzeng2014deep}. More recently, adversarial training methods have gained popularity for bridging domain gaps effectively~\citep{ajakan2014domain,ganin2015unsupervised,ganin2016domain,tzeng2017adversarial,isola2017image,lim2017geometric}. These methods leverage a discriminator to distinguish between domains, encouraging the generator to produce features or outputs that are indistinguishable across domains.

In this paper, we adopt a discriminator inspired by adversarial training-based approaches to reduce the domain gap in embedded features between multi-agent and single-agent datasets. This step ensures that the domain-adapted embeddings provide robust guidance for generation training on single-agent datasets in the second stage of our proposed method.

\nbf{Neural Radiance Fields}
Neural Radiance Fields (NeRF) have significantly advanced 2D novel view synthesis (NVS) by encoding scenes as implicit volumetric functions optimized through ray-marching~\citep{mildenhall2021nerf}. While effective in generating high-quality novel views, NeRF requires dense multi-view images and suffers from high computational costs~\citep{barron2021mip, barron2022mip}. Extensions such as Mip-NeRF~\citep{barron2021mip} improve aliasing, and depth-supervised variants reduce multi-view dependency~\citep{deng2022depth, roessle2022dense}.
In 3D LiDAR-based NVS, NeRF-inspired methods like LiDAR-NeRF~\citep{tao2024lidarnerf}, Neural LiDAR Fields~\citep{huang2023neural}, and NeRF-LiDAR~\citep{zhang2024nerf} adapt implicit representations to synthesize novel LiDAR views. These approaches enhance reconstruction but struggle with sparse data, large-scale outdoor scenes, and dynamic objects, as ray-marching is inefficient for LiDAR’s discrete nature~\citep{tao2024lidarnerf}.

Our work shares similarities with NeRF-based LiDAR generation~\citep{huang2023neural, tao2024lidarnerf, zhang2024nerf, zheng2024lidar4d} as we also synthesize LiDAR point clouds. However, unlike these methods focused on scene reconstruction from multiple samples (\eg, views, time frames), \ours generates collaborative driving data from a single frame. Instead of modeling implicit densities, \ours directly generates LiDAR point clouds with spatial consistency even at a long distance, enabling single-agent datasets to be converted into multi-agent data for autonomous driving.

\begin{table*}[t]
\centering
\resizebox{.99\linewidth}{!}{
\setlength{\tabcolsep}{4px}
\begin{tabular}{llcccccccccccccccc}
\toprule
\multirow{2}{*}{Method} & \multirow{2}{*}{Train Data} & \multicolumn{4}{c}{0 Add. Scene} & \multicolumn{4}{c}{5 Add. Scene} & \multicolumn{4}{c}{10 Add. Scene} & \multicolumn{4}{c}{22 Add. Scene} \\ \cmidrule(lr){3-6} \cmidrule(lr){7-10} \cmidrule(lr){11-14} \cmidrule(lr){15-18}
& & s & m & l & all & s & m & l & all & s & m & l & all & s & m & l & all \\ \midrule

No Fusion & ego's gt only & 0.67 & 0.41 & 0.13 & 0.40 & 0.85 & 0.66 & 0.22 & 0.57 & 0.82 & 0.60 & 0.20 & 0.53 & 0.87 & 0.71 & 0.25 & 0.60 \\
\cdashline{1-18} \noalign{\vskip2pt}

& baseline & & & &  
& 0.01 & 0.00 & 0.00 & 0.01 & 0.38 & 0.06 & 0.04 & 0.16 & 0.87 & 0.59 & 0.41 & 0.61 \\
& ~~+gt~~~~~~(\textbf{oracle}) & 0.76 & 0.42 & 0.31 & 0.49 & 0.89 & 0.65 & 0.48 & 0.66 & 0.88 & 0.63 & 0.48 & 0.65 & 0.96 & 0.82 & 0.62 & 0.78 \\
 \rowcolor{lightcyan}
\cellcolor{white}\multirow{-3}{*}{Early Fushion~\citep{chen2019cooper}} & ~~+\ours (\textbf{ours}) & 0.75 & 0.38 & 0.29 & 0.46 & 0.87 & 0.63 & 0.46 & 0.65 & 0.89 & 0.66 & 0.50 & 0.67 & 0.96 & 0.80 & 0.63 & 0.78 \\ 

 \cdashline{1-18} \noalign{\vskip2pt}
& baseline &  &  & &  & 0.15 & 0.10 & 0.03 & 0.09 & 0.54 & 0.33 & 0.19 & 0.35 & 0.91 & 0.77 & 0.50 & 0.71 \\
& ~~+gt~~~~~~(\textbf{oracle}) & 0.74 & 0.57 & 0.36 & 0.55 & 0.93 & 0.82 & 0.52 & 0.74 & 0.89 & 0.75 & 0.49 & 0.70 & 0.95 & 0.85 & 0.56 & 0.77 \\
 \rowcolor{lightcyan}
\cellcolor{white}\multirow{-3}{*}{Late Fushion~\citep{xu2022opv2v}} & ~~+\ours (\textbf{ours}) & 0.71 & 0.49 & 0.32 & 0.50 & 0.90 & 0.75 & 0.47 & 0.69 & 0.84 & 0.68 & 0.43 & 0.63 & 0.95 & 0.84 & 0.53 & 0.75 \\ 

\cdashline{1-18} \noalign{\vskip2pt}
& baseline &  &  &  &  & 0.66 & 0.31 & 0.28 & 0.41 & 0.86 & 0.60 & 0.48 & 0.64 & 0.94 & 0.76 & 0.56 & 0.74 \\
& ~~+gt~~~~~~(\textbf{oracle}) & 0.94 & 0.80 & 0.62 & 0.77 & 0.96 & 0.84 & 0.69 & 0.82 & 0.96 & 0.82 & 0.67 & 0.79 & 0.98 & 0.87 & 0.69 & 0.82 \\
 \rowcolor{lightcyan}
\cellcolor{white}\multirow{-3}{*}{AttFuse~\citep{xu2022opv2v}} & ~~+\ours (\textbf{ours}) & 0.90 & 0.73 & 0.56 & 0.72 & 0.95 & 0.81 & 0.65 & 0.79 & 0.94 & 0.79 & 0.62 & 0.77 & 0.98 & 0.88 & 0.75 & 0.86 \\

\cdashline{1-18} \noalign{\vskip2pt}
& baseline &  &  &  &  & 0.66 & 0.44 & 0.28 & 0.48 & 0.83 & 0.53 & 0.37 & 0.60 & 0.91 & 0.68 & 0.46 & 0.70 \\
& ~~+gt~~~~~~(\textbf{oracle}) & 0.87 & 0.71 & 0.50 & 0.71 & 0.88 & 0.73 & 0.57 & 0.74 & 0.91 & 0.77 & 0.63 & 0.78 & 0.94 & 0.80 & 0.66 & 0.81 \\
 \rowcolor{lightcyan}
\cellcolor{white}\multirow{-3}{*}{V2X-ViT~\citep{xu2022v2xvit}} & ~~+\ours (\textbf{ours}) & 0.84 & 0.65 & 0.40 & 0.65 & 0.88 & 0.72 & 0.50 & 0.71 & 0.90 & 0.74 & 0.55 & 0.74 & 0.94 & 0.79 & 0.59 & 0.78 \\ \hline
\bottomrule
\end{tabular}
}
\centering
\captionsetup{width=1.\linewidth}
\caption{
\label{tab:opv2v_full}
\small \textbf{Results on OPV2V with limited labeled data.} Using generated point clouds consistently achieves results comparable to oracles, demonstrating the quality of the generation. With additional labeled scenes, the gap is further minimized.}
\end{table*}

\section{Existing Datasets for CAV}
\label{sup:datasets}

We summarize existing CAV datasets in \cref{tab:datasets}, highlighting the current state of CAV research. At the time of this paper, no real-world dataset includes both dynamic and static agents and supports more than two agents, primarily due to the challenges of real-world data collection. Additionally, some datasets are limited to vehicle-only labels or a single sensor modality. These limitations drive our work, pushing boundaries and introducing a new research direction.

As shown in \cref{fig:teaser} and \cref{fig:teaser2}, \ours demonstrates strong potential to scale up the number of agents---both static and dynamic---through the proposed generation framework. Empirical results in \cref{tab:ft} further validate that the generated point clouds can enhance CAV development.

\begin{figure}[t]
\centering
\includegraphics[trim={0cm, 0cm, 0cm, 0cm},clip,width=1\linewidth]{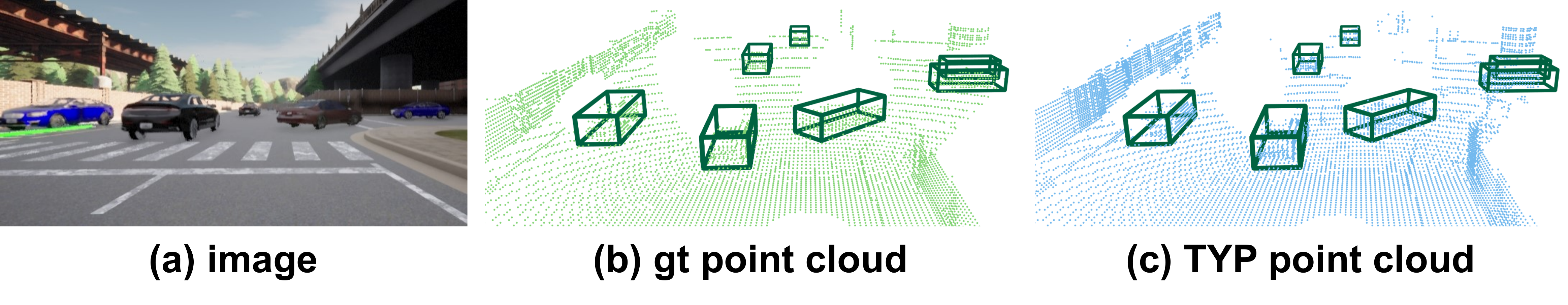}
\caption{
\label{fig:gt}
\small \textbf{Visualization with validation data of OPV2V.} The generated point clouds are well-aligned with the ground-truth bounding boxes and follow the physics (\eg, occluded areas).
}
\end{figure}

\section{Results on OPV2V}
\label{sup:opv2v}

In \cref{tab:opv2v} of the main paper, we validate the quality of the generated point clouds by replacing the ground-truth point clouds of the reference agents with the generated ones on the OPV2V dataset~\citep{xu2022opv2v}. In this \supp, we investigate the impact of having access to a limited amount of labeled data.

\nbf{Setting} As outlined in \cref{exp:opv2v} of the main paper, the original training set of 44 scenes was split into two halves: the first 22 scenes were used to train the generation model, while the remaining 22 scenes were used for inference to generate point clouds of reference agents. Here, we further utilize the first split as a source of limited labeled data.  

\nbf{Results} 
Firstly, the results in \cref{tab:opv2v_full} exhibit a consistent trend with \cref{tab:opv2v} in the main paper, demonstrating that using generated point clouds achieves results comparable to those obtained with ground-truth point clouds (oracle).
Secondly, the results in \cref{tab:opv2v_full} highlight that incorporating additional limited labeled data further reduces the gap between using ground-truth and generated point clouds. For example, in Early Fusion with 22 additional labeled scenes, the performance with generated point clouds matches that of ground-truth point clouds (\ie, both achieve $0.78$).

\section{Statistical Results of Experiments}
\label{sup:stat}
\begin{table*}[ht]
\centering
\resizebox{.75\linewidth}{!}{
\setlength{\tabcolsep}{6px}
\begin{tabular}{llcccc}
\toprule
\multirow{2}{*}{Method} & \multirow{2}{*}{Train Data} & \multicolumn{4}{c}{Average (\textit{$\pm$ std.})} \\ \cmidrule(lr){3-6} 
& & s & m & l & all \\ \midrule

& gt~~~~~~(\textbf{oracle}) & 0.78 (\textit{$\pm$ 0.02}) & 0.44 (\textit{$\pm$ 0.04}) & 0.35 (\textit{$\pm$ 0.04}) & 0.52 (\textit{$\pm$ 0.02}) \\
& \ours (\textbf{single}) & 0.63 (\textit{$\pm$ 0.13}) & 0.34 (\textit{$\pm$ 0.15}) & 0.24 (\textit{$\pm$ 0.09}) & 0.40 (\textit{$\pm$ 0.11}) \\
\rowcolor{lightcyan} 
\cellcolor{white}\multirow{-3}{*}{Early Fusion~\citep{chen2019cooper}} 
& \ours (\textbf{ours}) & 0.75 (\textit{$\pm$ 0.04}) & 0.40 (\textit{$\pm$ 0.09}) & 0.30 (\textit{$\pm$ 0.08}) & 0.47 (\textit{$\pm$ 0.06}) \\

\cdashline{1-6} 
\noalign{\vskip2pt} 
& gt~~~~~~(\textbf{oracle}) & 0.77 (\textit{$\pm$ 0.07}) & 0.61 (\textit{$\pm$ 0.10}) & 0.40 (\textit{$\pm$ 0.06}) & 0.58 (\textit{$\pm$ 0.08}) \\
& \ours (\textbf{single}) & 0.75 (\textit{$\pm$ 0.03}) & 0.55 (\textit{$\pm$ 0.05}) & 0.35 (\textit{$\pm$ 0.04}) & 0.55 (\textit{$\pm$ 0.04}) \\
\rowcolor{lightcyan} 
\cellcolor{white}\multirow{-3}{*}{Late Fusion~\citep{xu2022opv2v}} 
& \ours (\textbf{ours}) & 0.79 (\textit{$\pm$ 0.07}) & 0.60 (\textit{$\pm$ 0.10}) & 0.37 (\textit{$\pm$ 0.05}) & 0.58 (\textit{$\pm$ 0.07}) \\ 
 
\cdashline{1-6}
\noalign{\vskip2pt} 
& gt~~~~~~(\textbf{oracle}) & 0.93 (\textit{$\pm$ 0.01}) & 0.78 (\textit{$\pm$ 0.02}) & 0.62 (\textit{$\pm$ 0.02}) & 0.77 (\textit{$\pm$ 0.02}) \\
& \ours (\textbf{single}) & 0.90 (\textit{$\pm$ 0.02}) & 0.70 (\textit{$\pm$ 0.03}) & 0.54 (\textit{$\pm$ 0.02}) & 0.70 (\textit{$\pm$ 0.02}) \\
\rowcolor{lightcyan} 
\cellcolor{white}\multirow{-3}{*}{AttFuse~\citep{xu2022opv2v}} 
& \ours (\textbf{ours}) & 0.91 (\textit{$\pm$ 0.00}) & 0.72 (\textit{$\pm$ 0.02}) & 0.56 (\textit{$\pm$ 0.01}) & 0.72 (\textit{$\pm$ 0.00}) \\ 
\hline
\bottomrule
\end{tabular}
}
\centering
\captionsetup{width=1.\linewidth}
\caption{
\label{tab:opv2v_mean}
\small \textbf{Statistical Results on OPV2V.} We report the mean and standard deviation from multiple runs of the same experiment, demonstrating the consistency of the results. Additionally, we include the performance of point clouds generated using single-stage training, which is consistently worse than the two-stage approach, highlighting the generation quality of two-stage training.}
\end{table*}

In \cref{tab:opv2v} of the main paper, we validate the quality of the generated point clouds by replacing the ground-truth point clouds of the reference agents with the generated ones on the OPV2V dataset~\citep{xu2022opv2v}. In this \supp, we extend this evaluation by conducting two additional runs (\ie, three in total) and present the statistical results in \cref{tab:opv2v_mean}. The results consistently demonstrate that using the generated point clouds achieves performance comparable to that of the ground-truth (oracle) point clouds, highlighting the robustness, consistency, and reproducibility of our approach.

\section{Single-Stage \vs Multi-Stage Training}
\label{sup:stage}
In \cref{tab:step} of the main paper, we compare the quality of generated point clouds between single-stage and the proposed multi-stage training by evaluating the distance between generated and ground-truth samples. In this \supp, we extend this analysis by conducting CAV training using point clouds generated by the single-stage training model. 

The results in \cref{tab:opv2v_mean} demonstrate that the performance of single-stage training consistently lags behind the proposed multi-stage approach, particularly in scenarios that directly rely on point clouds (\ie, early fusion). Furthermore, the multi-stage method remains essential for translating single-agent datasets into collaborative versions, underscoring its critical role in the proposed framework (\cf \cref{sec:adapt,exp:colwaymo} in the main paper).

\section{More Qualitative Results}
\label{sup:qual}
We present additional examples of \ours in \cref{fig:teaser2}. These examples demonstrate the ability to designate \emph{any location} as a reference, effectively simulating both static and dynamic agents communicating with the ego vehicle. This flexibility overcomes the limitations of existing real-world CAV datasets, which are often constrained by specific communication types (\ie, vehicle-to-vehicle or vehicle-to-infrastructure) and a limited number of agents. Furthermore, the generated point clouds are both realistic and semantically consistent with the ego agent's perception.

In \cref{fig:colwaymo2}, we provide more examples of the collaborative version of the Waymo dataset (\ie, ColWaymo), which was utilized to pre-train the detector for CAV tasks, as discussed in \cref{exp:colwaymo,tab:ft} of the main paper. These examples further highlight the high quality of the point clouds generated by \ours and underscore its potential to significantly scale up datasets for CAV research.

\begin{figure*}
\centering
\includegraphics[trim={0cm, 0cm, 0cm, 0cm},clip,width=.9\linewidth]{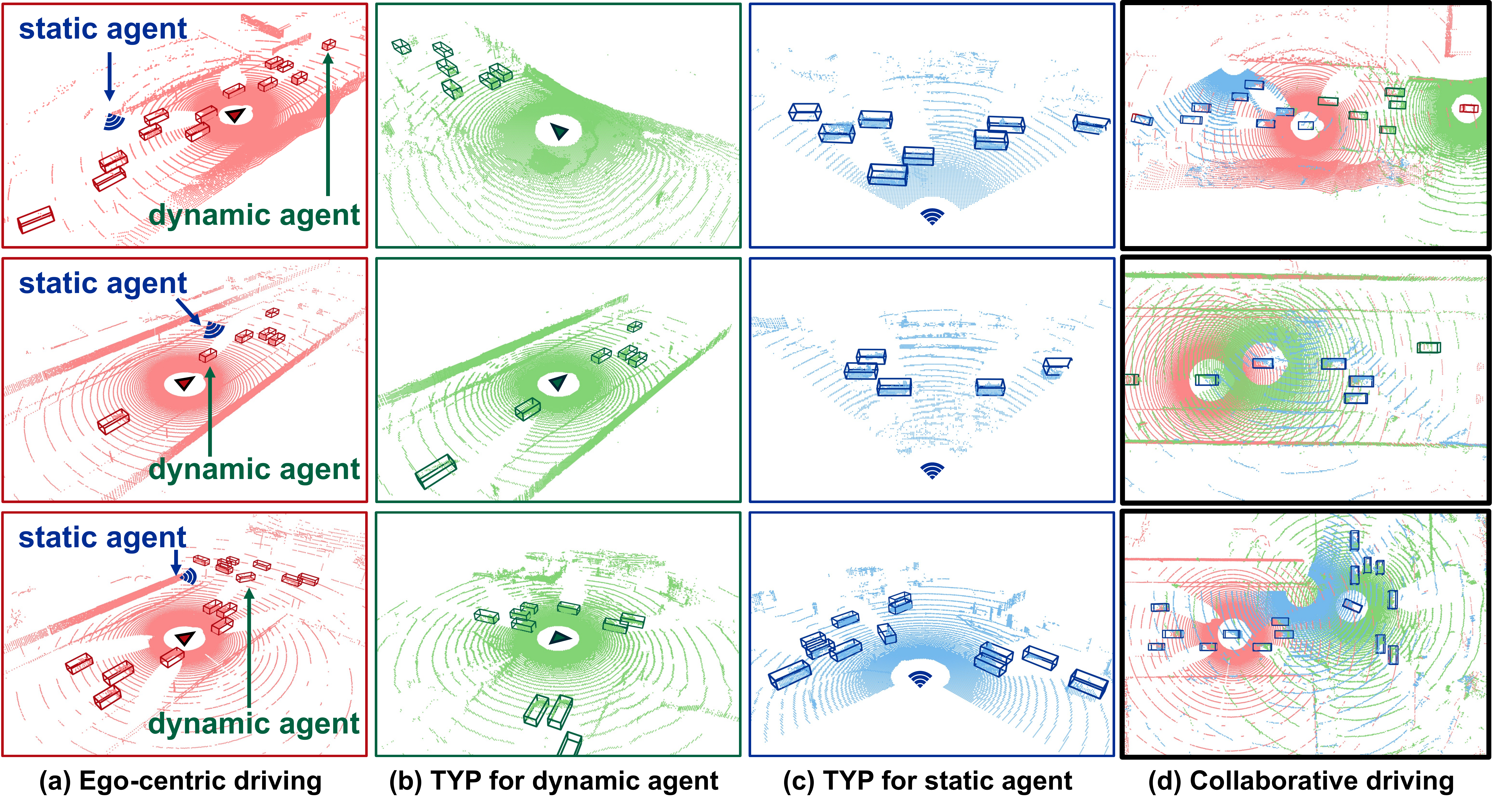}

\caption{
\label{fig:teaser2}
\small \textbf{Illustration of the proposed problem and solution, Transfer Your Perspective (\ours).} (a) A given sensory data captured by the {\color{red} ego-car (red triangle)}. (b) A generated sensory data by \ours, seeing from the viewpoint of \textcolor{customgreen}{another vehicle (green triangle)} in the same scene. (c) A generated sensory data, seeing from an imaginary {\color{blue} static agent like roadside units (blue icon)}. (d) Putting all the sensory data together, given or generated, \ours enables the development of collaborative perception with little or no real collaborative driving data.}
\end{figure*}

\begin{figure*}
\centering
\includegraphics[trim={0cm, 0cm, 0cm, 0cm},clip,width=.9\linewidth]{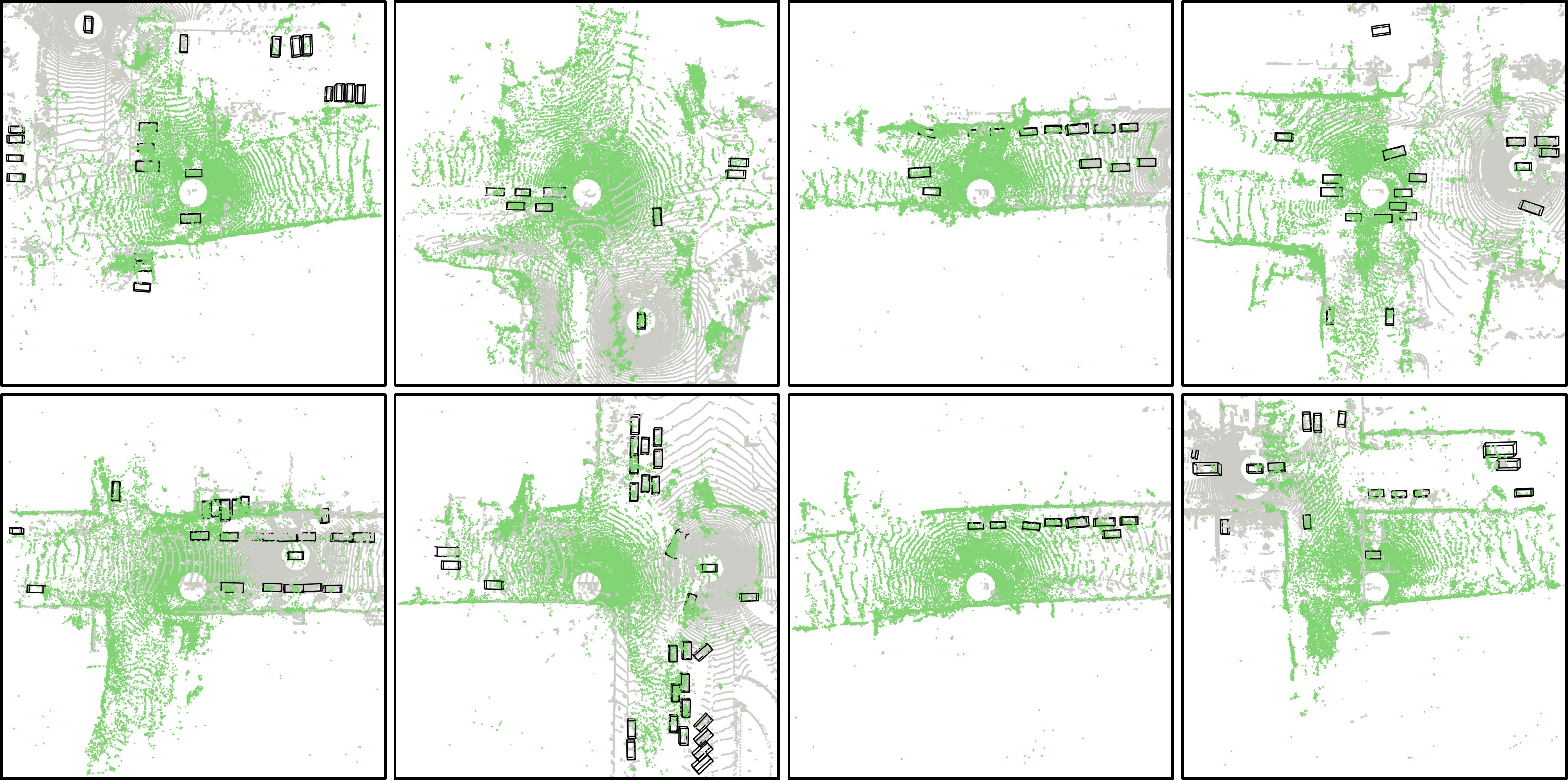}
\caption{
\label{fig:colwaymo2}
\small \textbf{Qualitative results on Collaborative Waymo.} The gray point clouds are from the original single-agent dataset and the green are generated by \ours conditioning on them.}

\end{figure*}

\section{Future Work}
\label{sup:future}

This paper follows the existing benchmark~\citep{xu2022opv2v, xu2023v2v4real} to focus on vehicle-like objects. However, \ours is scalable and can extend to broader object categories when semantic information is available (\cf \cref{sec:setup} and \cref{sec:input}). We are also open to exploring cross-modality generation in future research.
 \fi

\end{document}